\documentclass[letterpaper]{article} 
\usepackage{aaai2026}  
\usepackage{times}  
\usepackage{helvet}  
\usepackage{courier}  
\usepackage[hyphens]{url}  
\usepackage{graphicx} 
\usepackage{natbib}  
\usepackage{caption} 
\usepackage{algorithm}
\usepackage{algorithmic}
\usepackage{amsmath}
\usepackage{amssymb}
\usepackage{amsthm, bm}
\usepackage{booktabs}
\usepackage{multirow}

\usepackage{newfloat}
\usepackage{listings}
\DeclareCaptionStyle{ruled}{labelfont=normalfont,labelsep=colon,strut=off} 
\floatstyle{ruled}
\newfloat{listing}{tb}{lst}{}
\floatname{listing}{Listing}
\title{MAPS: \underline{M}ulti-\underline{A}gent \underline{P}ersonality \underline{S}haping for Collaborative Reasoning}

\author{
    Jian Zhang\equalcontrib\textsuperscript{\rm 1,2},
    Zhiyuan Wang\equalcontrib\textsuperscript{\rm 1,3},
    Zhangqi Wang\equalcontrib\textsuperscript{\rm 1,3},
    Fangzhi Xu\textsuperscript{\rm 1,3},\\
    Qika Lin\textsuperscript{\rm 4},
    Lingling Zhang\textsuperscript{\rm 1,3},
    Rui Mao\textsuperscript{\rm 5},
    Erik Cambria\textsuperscript{\rm 5},
    Jun Liu\textsuperscript{\rm 1,3}\thanks{Corresponding Author}
}
\affiliations{
    \textsuperscript{\rm 1}School of Computer Science and Technology, Xi’an Jiaotong University, China\\
    \textsuperscript{\rm 2}MOE KLINNS Lab, Xi’an Jiaotong University, China\\
    \textsuperscript{\rm 3}Shaanxi Province Key Laboratory of Big Data Knowledge Engineering, China\\
    \textsuperscript{\rm 4}Saw Swee Hock School of Public Health, National University of Singapore, Singapore\\
    \textsuperscript{\rm 5}College of Computing and Data Science, Nanyang Technological University, Singapore\\
    zhangjian062422@stu.xjtu.edu.cn, liukeen@xjtu.edu.cn
}

\usepackage{bibentry}

\begin{document}

\maketitle

\begin{abstract}
Collaborative reasoning with multiple agents offers the potential for more robust and diverse problem-solving. However, existing approaches often suffer from homogeneous agent behaviors and lack of reflective and rethinking capabilities. We propose \underline{\textbf{M}}ulti-\underline{\textbf{A}}gent \underline{\textbf{P}}ersonality \underline{\textbf{S}}haping (\textbf{(MAPS}), a novel framework that enhances reasoning through agent diversity and internal critique. Inspired by the Big Five personality theory, MAPS assigns distinct personality traits to individual agents, shaping their reasoning styles and promoting heterogeneous collaboration. To enable deeper and more adaptive reasoning, MAPS introduces a \textit{Critic} agent that reflects on intermediate outputs, revisits flawed steps, and guides iterative refinement. This integration of personality-driven agent design and structured collaboration improves both reasoning depth and flexibility. Empirical evaluations across three benchmarks demonstrate the strong performance of MAPS, with further analysis confirming its generalizability across different large language models and validating the benefits of multi-agent collaboration.
\end{abstract}

\begin{links}
    \link{Code}{https://github.com/exoskeletonzj/MAPS}
\end{links}

\section{Introduction}
\label{sec:intro}

\begin{figure}[t]
 \includegraphics[width=\columnwidth]{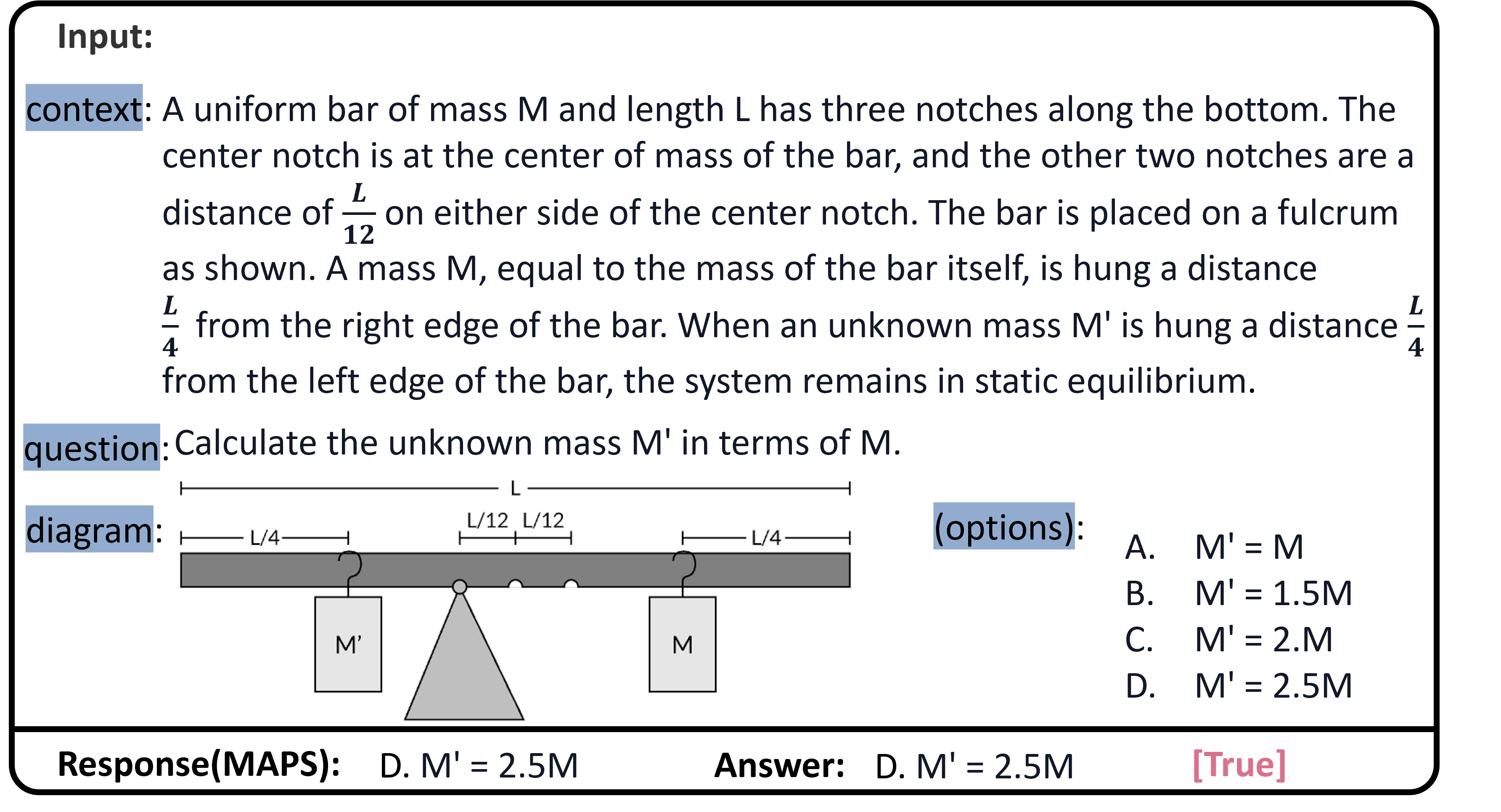}
 \caption{
 An example of a multimodal scientific multiple-choice problem.
 The correct answer is derived based on the reasoning over inputs that include context, question, and diagram.
 }
 \label{fig_example}
\end{figure}

Solving complex reasoning problems~\cite{bhattacharya2024large,li2024multimodal,he2024olympiadbench} often requires more than just accurate perception and factual recall---it demands nuanced interpretation, multi-step inference, and the ability to adapt when initial attempts fail~\cite{fu2024isobench,li2024survey}. While large language models (LLMs) demonstrate promising capabilities in solving such problems, they frequently fall short in scenarios requiring sustained reasoning, internal verification, and flexible strategy revision~\cite{anand2024mm,alasadi2024multimodal,gao2024cantor,wang2024t,xu2025large}. A key challenge lies in how to effectively combine diverse reasoning strategies and incorporate mechanisms for intermediate reflection. Prior work has explored both single-agent solutions and simple collaborative setups (e.g., paired discussion or voting)~\cite{kaesberg2025voting}, yet these approaches often suffer from rigid behaviors and limited capacity for self-correction. This raises an important research question: \emph{how can we enhance the depth and adaptability of reasoning by enabling more flexible and reflective solution processes?} Figure~\ref{fig_example} illustrates a representative scenario that embodies these challenges, where successful problem solving requires interpreting multimodal input and applying domain-specific reasoning.

\begin{figure*}[t]
	\large
	\centering
	\includegraphics[scale=0.4]{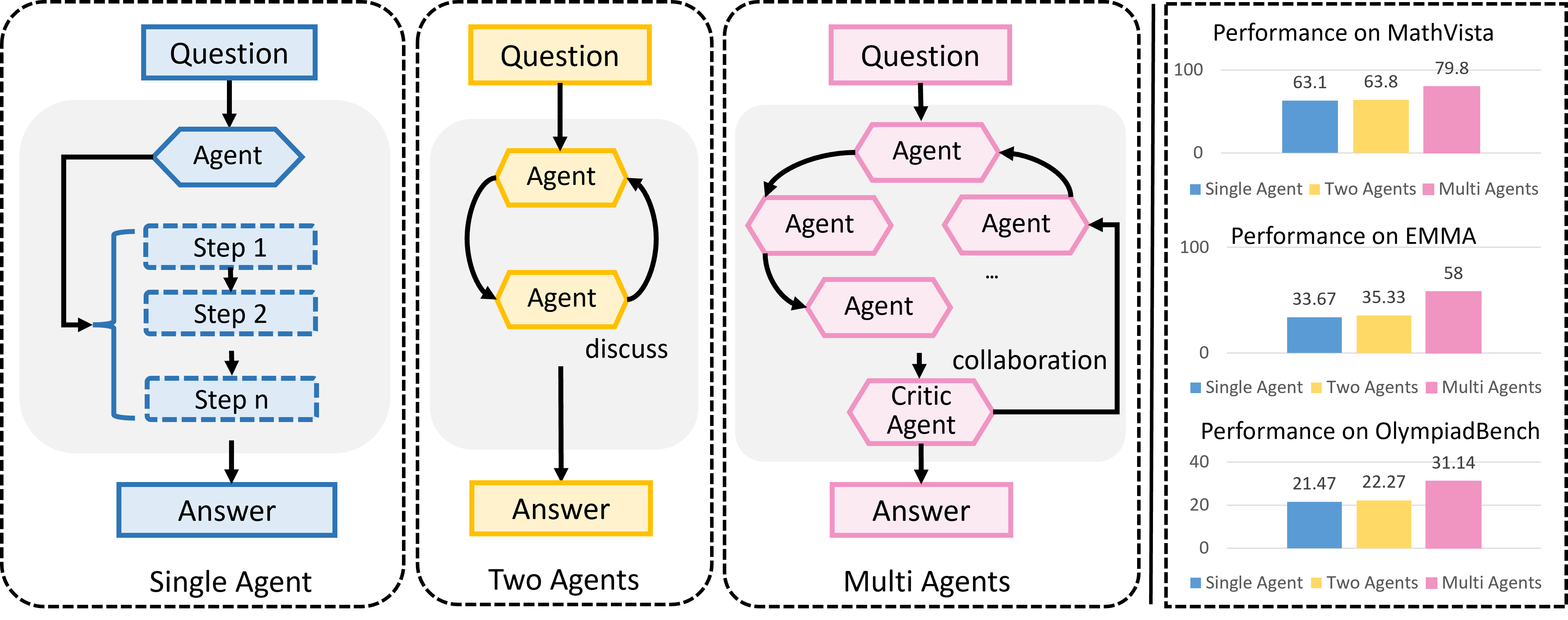}
	\caption{
Comparison of reasoning paradigms. Single-agent and two-agent approaches offer limited adaptability. MAPS enables dynamic collaborative reasoning. Right: Built on GPT-4o, MAPS achieves the best performance across three benchmarks.}
	\label{fig_compare}
\end{figure*}

As illustrated in Figure~\ref{fig_compare}, existing methods~\cite{wang2024examining,landau2024computational,hardiansyah2024analysis,zhang2024mm,caffagni2024revolution,qiu2024application} for complex reasoning problems often adopt single-agent solutions or simple two-agent collaborations. Although effective to some extent, these setups suffer from \textbf{homogeneous agent behaviors}—reasoning steps tend to repeat similar patterns, leading to redundancy and premature convergence. The lack of diversity limits exploration and reduces the chance of identifying alternative perspectives or correcting errors, especially in multi-turn settings where roles and strategies remain undifferentiated.

Another issue is the \textbf{lack of reflective and rethinking capabilities} in existing approaches. As shown in Figure~\ref{fig_compare}, the interaction between two agents is often linear and lacks mechanisms for feedback or revision. Even with multiple turns, agents rarely revisit earlier reasoning or correct initial misconceptions. In contrast, human reasoning is inherently iterative: People reflect, reassess, and adjust their thinking over time. Without structured reflection, current methods risk premature convergence and fail to recover from early-stage errors.

To address these two issues, we propose \textbf{MAPS} (\textbf{M}ulti-\textbf{A}gent \textbf{P}ersonality \textbf{S}haping), a collaborative reasoning framework that enhances both diversity and adaptability in complex problem solving. Inspired by the Big Five personality theory~\cite{almagor1995big,benet1995big,simms2007big}, MAPS shapes the reasoning behaviors of a set of role-specialized agents through distinct personality traits, promoting heterogeneous collaboration and mitigating behavioral homogeneity. To enable reflective thinking and iterative refinement, MAPS further introduces a \textit{Critic} agent that revisits intermediate outputs, identifies flawed steps, and provides structured feedback. This integration of personality-guided agent design and internal critique supports deeper, more flexible reasoning aligned with human cognitive processes.

\begin{figure}[t]
  \large
  \centering
 \includegraphics[width=0.6\columnwidth]{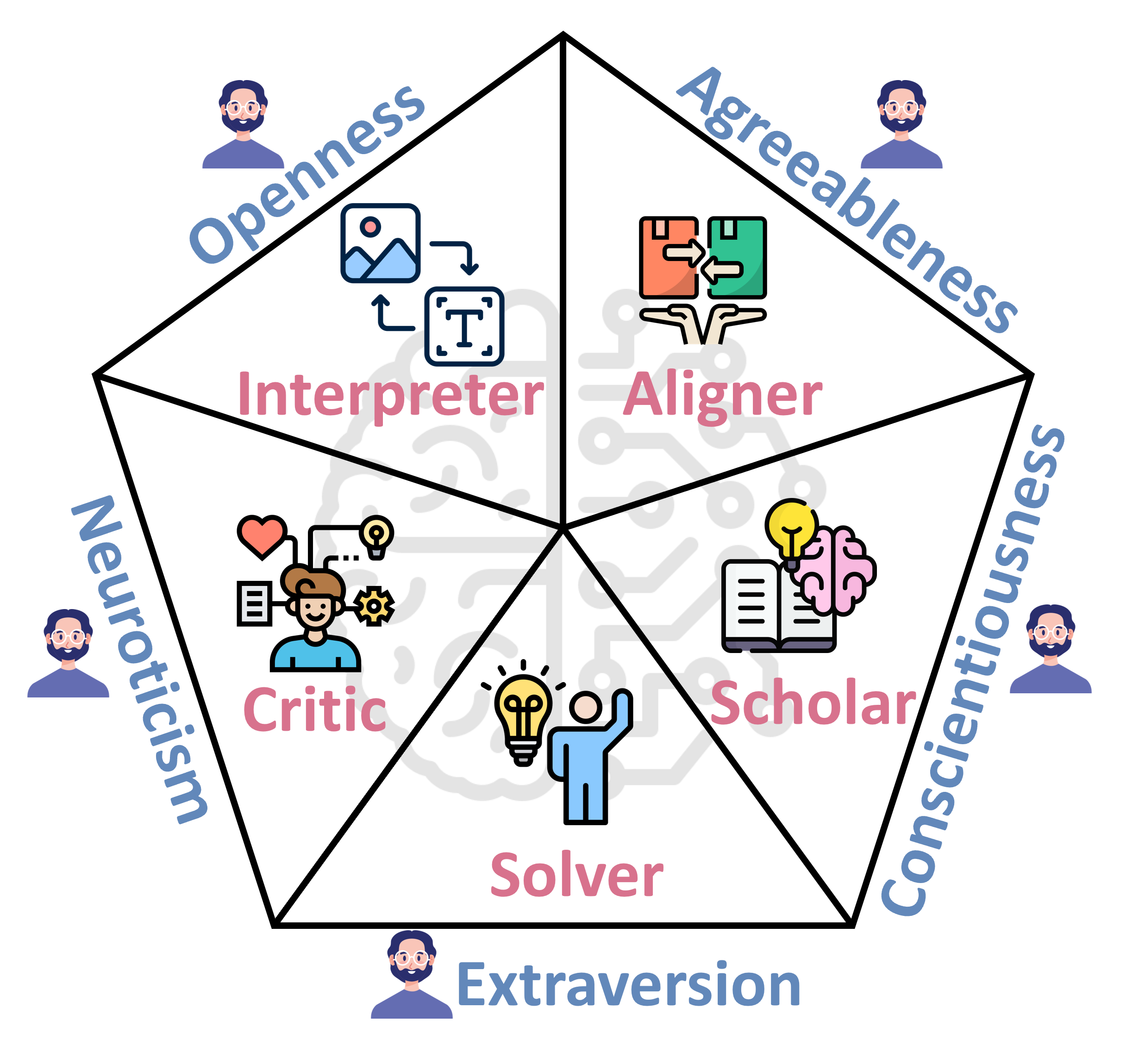}
 \caption{The corresponding relation between the Big Five Personality theory and the five function-specific agents.
 }
 \label{fig_example1}
\end{figure}

As shown in Figure~\ref{fig_example1}, 
the \textit{interpreter} (Openness) explores information from multiple angles, the \textit{aligner} (Agreeableness) reconciles visual and textual cues, the \textit{scholar} (Conscientiousness) ensures precision and factual grounding, the \textit{solver} (Extraversion) drives goal-oriented conclusions, and the \textit{critic} (Neuroticism) questions assumptions and detects flaws. The \textit{critic} is further inspired by Socratic questioning~\cite{elder1998role,paul2019thinker}, providing reflective feedback throughout the multi-stage reasoning process. Together, these roles enable a structured yet flexible collaboration that supports deeper and more reliable problem solving.

We conduct extensive experiments on three challenging benchmarks: MathVista~\cite{lu2023mathvista}, OlympiadBench~\cite{he2024olympiadbench}, and EMMA~\cite{hao2025can}. These datasets cover a wide range of complex reasoning tasks.
MAPS consistently outperforms baseline methods across all datasets, confirming its effectiveness in enhancing both reasoning depth and adaptability. We further evaluate MAPS under different base LLMs, and results consistently show its superiority across model backbones. Additional analyses examine the impact of the feedback mechanism and the overall efficiency of the framework.

\noindent\textbf{Our main contributions are as follows:}

$\bullet$ We propose \textbf{MAPS}, a multi-agent reasoning framework. To the best of our knowledge, this is the first work that incorporates personality shaping based on the Big Five theory into collaborative reasoning.

$\bullet$ MAPS addresses two key challenges in existing methods: behavioral homogeneity and lack of reflection. It assigns distinct personality traits to agents and introduces a \textit{Critic} inspired by Socratic questioning.

$\bullet$ We conduct extensive experiments on three scientific reasoning benchmarks. MAPS achieves consistent performance gains (up to 15.84\%) and generalizes well across different tasks and base models.

\section{Methodology}

\begin{figure*}[t]
	\large
	\centering
	\includegraphics[scale=0.35]{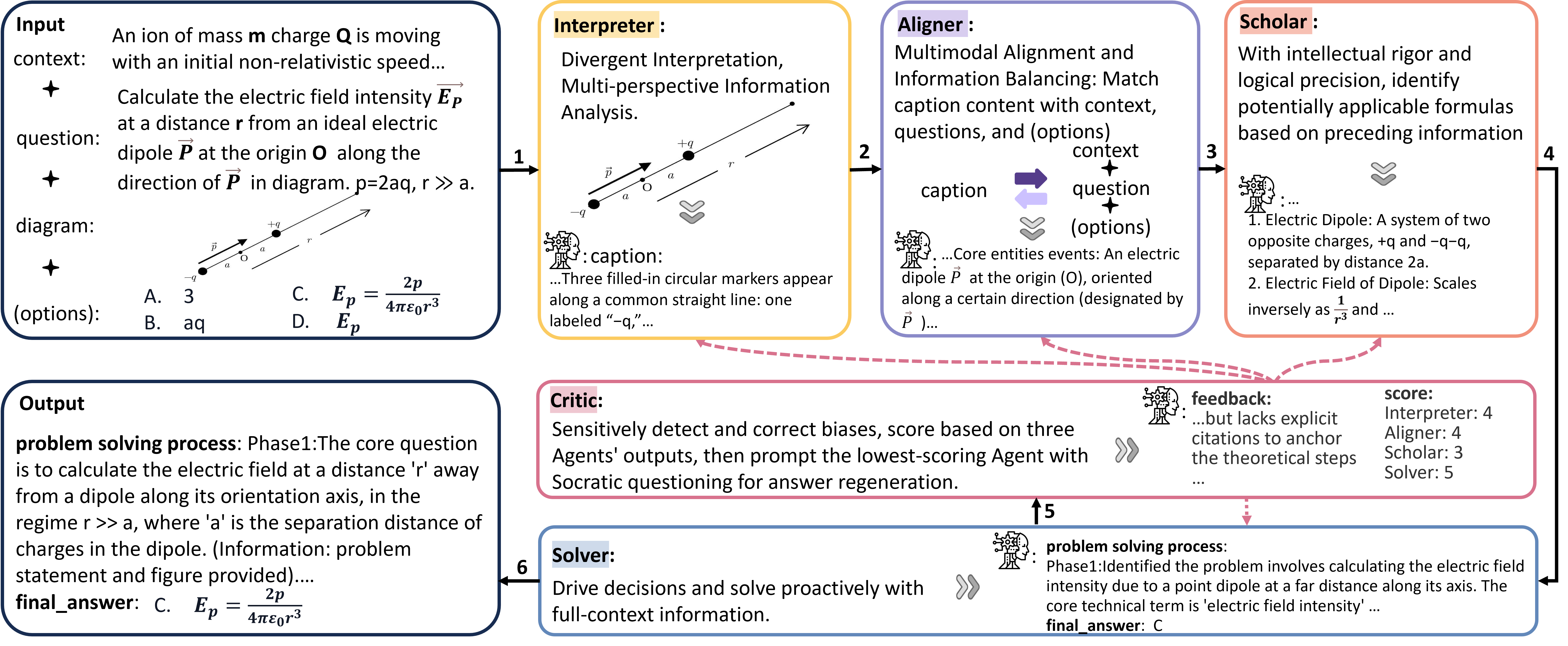}
	\caption{
  The overall architecture of MAPS. The framework consists of five functional agents inspired by the Big Five personality theory. The core reasoning process is carried out by four specialized agents—\textit{Interpreter}, \textit{Aligner}, \textit{Scholar}, and \textit{Solver}—each responsible for a distinct stage in solving complex reasoning problems. Finally, the \textit{Critic} agent provides reflective feedback and correction to enhance accuracy and interpretability.
}
	\label{fig_model}
\end{figure*}

This section introduces MAPS in four key components: \textit{Preliminaries}, \textit{Agentic Interaction Logic}, the \textit{Four-Step Reasoning} process, and the \textit{Critic and Feedback} mechanism.

\subsection{Preliminaries}

\paragraph{Task Definition.}
We define the complex reasoning task as a mapping from structured multimodal inputs to a target answer space. Let \(\mathcal{D}\), \(\mathcal{C}\), and \(\mathcal{Q}\) denote the input spaces of diagrams, contexts, and questions, respectively, and let \(\mathcal{A}\) denote the answer space. Each instance is a triplet \((d_i, c_i, q_i) \in \mathcal{D} \times \mathcal{C} \times \mathcal{Q}\), and the goal is to predict the corresponding answer \(a_i \in \mathcal{A}\).

The task is thus modeled as a function:
\begin{equation}
\mathcal{M}: \mathcal{D} \times \mathcal{C} \times \mathcal{Q} \rightarrow \mathcal{A}, \quad a_i = \mathcal{M}(d_i, c_i, q_i),
\end{equation}
where \(\mathcal{M}\) denotes the reasoning system responsible for processing visual-textual input and generating the output.

\paragraph{Agent-Based Modeling.}
In MAPS, the reasoning process \(\mathcal{M}\) is decomposed into a sequence of collaborative stages executed by a set of role-specialized agents \(\{\mathcal{A}_1, \mathcal{A}_2, \dots, \mathcal{A}_K\}\), where each agent \(\mathcal{A}_k\) performs a specific function conditioned on both the original input and intermediate reasoning states.

Let \(\mathcal{S}_0 = (d_i, c_i, q_i)\) be the initial input state. The overall reasoning unfolds as a staged transformation:
\begin{equation}
\mathcal{S}_{k} = \mathcal{A}_k(\mathcal{S}_{k-1}), \quad \text{for } k = 1, 2, \dots, K,
\end{equation}
where \(\mathcal{S}_k\) denotes the intermediate reasoning state after the \(k\)-th agent’s operation. The final answer is extracted as \(a_i = \mathrm{Extract}(\mathcal{S}_K)\).
This formulation reflects the staged reasoning in MAPS, where specialized agents collaboratively refine the solution.

\subsection{Agents Interaction Logic} \label{2.2}

As illustrated in Algorithm~\ref{alg:training}, this section introduces the interaction logic among the five personality-driven agents in MAPS. Complex problem reasoning requires multimodal semantic integration and structured inference over multiple steps. MAPS models this process as a functional composition of role-specialized reasoning agents, each shaped by a distinct personality trait.

Let \(\mathbf{x} = (d_i, c_i, q_i)\) be the input, and let \(\mathbf{p}_k \in \mathbb{R}^m\) denote the personality embedding associated with agent \(\mathcal{A}_k\). The entire collaborative reasoning process is represented as:

\begin{equation}
a_i = \mathcal{F}(\mathbf{x}; \mathbf{p}_1, \dots, \mathbf{p}_4) = \mathcal{A}_4 \circ \mathcal{A}_3 \circ \mathcal{A}_2 \circ \mathcal{A}_1(\mathbf{x}),
\end{equation}

where each agent \(\mathcal{A}_k(\cdot; \mathbf{p}_k)\) executes its stage conditioned on its personality vector \(\mathbf{p}_k\), contributing a unique reasoning perspective.

After the initial reasoning trajectory \(\mathcal{T} = \{\mathcal{S}_1, \mathcal{S}_2, \mathcal{S}_3, \mathcal{S}_4\}\) is generated, the \textit{Critic} agent performs a reflective evaluation by applying a feedback function:

\begin{equation}
\mathbf{f} = \texttt{Reflect}(\mathcal{T}), \quad \mathbf{f} = \{f_1, f_2, f_3, f_4\} \in \mathbb{R}^4,
\end{equation}

where each score \(f_k\) reflects the confidence of the \(k\)-th stage. If any \(f_k < \tau\), the corresponding stage is revised. This feedback loop complements the forward reasoning path, allowing MAPS to emulate deliberative human cognition through structured collaboration and critique.

\textit{\textbf{Proposition 1 (Monotonic Free-Energy Descent).}  
For the free energy
\(F^{(t)} = \mathbb{E}_{q^{(t)}}[-\!\log p(\mathbf{x},a_i\mid\theta)]
           + \mathrm{KL}\bigl(q^{(t)}\|\;p\bigr)\),
each Critic-triggered update satisfies}
\begin{equation}
F^{(t+1)} \;\le\; F^{(t)} .
\end{equation}
\textit{Thus the MAPS iteration produces a non-increasing free-energy sequence that converges to a stationary point.}
The full proof is provided in Appendix A.1.

\begin{algorithm}[ht]
\caption{MAPS Collaborative Reasoning with Reflective Feedback}
\label{alg:training}
\begin{algorithmic}[1]
\STATE \textbf{Input:} Diagram \(d_i\), Context \(c_i\), Question \(q_i\), Personality vectors \(\{\mathbf{p}_1,\dots,\mathbf{p}_4\}\)
\STATE \textbf{Initialize:} \(\mathcal{S}_0 = (d_i, c_i, q_i),\quad t \leftarrow 0\)
\REPEAT
    \STATE \textbf{\textit{Interpreter}}: \quad \(p_i = \mathcal{A}_1(\mathcal{S}_0;\mathbf{p}_1)\)
    \STATE \textbf{\textit{Aligner}}: \quad \(l_i = \mathcal{A}_2(p_i, c_i, q_i;\mathbf{p}_2)\)
    \STATE \textbf{\textit{Scholar}}: \quad \(s_i = \mathcal{A}_3(l_i, p_i, c_i, q_i;\mathbf{p}_3)\)
    \STATE \textbf{\textit{Solver}}: \quad \(a_i = \mathcal{A}_4(s_i, l_i, p_i;\mathbf{p}_4)\)
    \STATE \textbf{Reasoning Trajectory:} \quad \(\mathcal{T}^{(t)} = \{p_i, l_i, s_i, a_i\}\)
    \STATE \textbf{\textit{Critic}}: \quad \(\mathbf{f}^{(t)} = \texttt{Reflect}(\mathcal{T}^{(t)}),\quad f_k \in [0,1]\)
    \IF{all \(f_k \ge \tau\)}
        \STATE \textbf{break}
    \ELSE
        \STATE Identify stage \(k^* = \arg\min_k f_k\)
        \STATE Rerun agent \(\mathcal{A}_{k^*}\) with updated input
        \STATE \(t \leftarrow t + 1\)
    \ENDIF
\UNTIL{convergence}
\RETURN Final answer \(a_i\)
\end{algorithmic}
\end{algorithm}

\subsection{Four-Step Reasoning} \label{2.3}

Given input \(\mathbf{x} = (d_i, c_i, q_i)\), MAPS conducts a structured reasoning process across four personality-driven agents: \textit{Interpreter}, \textit{Aligner}, \textit{Scholar}, and \textit{Solver}. Each agent \(\mathcal{A}_k\) is parameterized by a personality embedding \(\mathbf{p}_k \in \mathbb{R}^m\), shaping its reasoning style and focus. The entire inference pipeline can be expressed as a function composition:

\begin{equation}
a_i = \mathcal{A}_4 \circ \mathcal{A}_3 \circ \mathcal{A}_2 \circ \mathcal{A}_1(\mathbf{x}; \mathbf{p}_1, \mathbf{p}_2, \mathbf{p}_3, \mathbf{p}_4),
\end{equation}

where each intermediate state \(\mathcal{S}_k\) is defined recursively as \(\mathcal{S}_k = \mathcal{A}_k(\mathcal{S}_{k-1}; \mathbf{p}_k)\), with \(\mathcal{S}_0 = \mathbf{x}\).

\noindent\textbf{\textit{Interpreter}}.  
The \textit{Interpreter} agent aims to extract structured visual semantics from diagram \(d_i\), translating them into a caption representation \(p_i\) that can be consumed by downstream language agents. Let \(\phi_{\text{vis}}(d_i)\) be a visual encoder, and \(\psi_{\text{lang}}(\cdot)\) a caption generator. Then the agent performs:

\begin{equation}
p_i = \mathcal{A}_1(d_i; \mathbf{p}_1) = \psi_{\text{lang}}\big(\phi_{\text{vis}}(d_i) + W_1 \mathbf{p}_1 \big),
\end{equation}

where \(W_1 \in \mathbb{R}^{d \times m}\) projects the personality embedding into the encoder space, modulating its attention to attributes like spatiality, color, or topology.

\noindent\textbf{\textit{Aligner}}.  
To resolve semantic mismatches across modalities, the \textit{Aligner} fuses the interpreted diagram caption \(p_i\) with the textual context \(c_i\) and question \(q_i\). The process outputs an alignment representation \(l_i\), optimized to preserve shared semantics and suppress modality conflict. Formally,

\begin{equation}
l_i = \mathcal{A}_2(p_i, c_i, q_i; \mathbf{p}_2) = \mathrm{CrossFuse}\big(p_i, c_i, q_i; \mathbf{p}_2\big),
\end{equation}

where \(\mathrm{CrossFuse}(\cdot)\) denotes a multi-head attention-based fusion operator, adaptively weighted by \(\mathbf{p}_2\) to emphasize visual or linguistic cues based on agent bias.

\noindent\textbf{\textit{Scholar}}.  
While \(l_i\) captures semantic consistency, complex reasoning often requires external knowledge supplementation. The \textit{Scholar} agent retrieves and integrates domain-specific knowledge \(\mathcal{K}(l_i)\), such as physics principles or mathematical theorems. We define:

\begin{equation}
s_i = \mathcal{A}_3(l_i, p_i, c_i, q_i; \mathbf{p}_3) 
= \mathrm{KnowAug}\big(l_i, \mathcal{K}(l_i); \mathbf{p}_3\big),
\end{equation}

where \(\mathrm{KnowAug}(\cdot)\) augments contextual embeddings with retrieved tuples \(\mathcal{K}(l_i)\) from a structured knowledge memory, and \(\mathbf{p}_3\) biases the agent toward formal rigor or heuristic reasoning.

\noindent\textbf{\textit{Solver}}.  
The \textit{Solver} agent aggregates all upstream outputs and executes logical composition to generate the final answer \(a_i\). Let \(\mathcal{H}_i = \{p_i, l_i, s_i\}\) be the hybrid reasoning state. The solver computes:

\begin{equation}
a_i = \mathcal{A}_4(\mathcal{H}_i; \mathbf{p}_4) 
= \mathrm{Deduct}\big(p_i, l_i, s_i; \mathbf{p}_4\big),
\end{equation}

where \(\mathrm{Deduct}(\cdot)\) is a constrained generation module that synthesizes the inputs under logical, numerical, or symbolic rules. The final prediction \(a_i \in \mathcal{A}\) may be a selected option or a free-form answer.

\noindent
Together, these four stages construct a trajectory \(\mathcal{T} = \{p_i, l_i, s_i, a_i\}\), on which the \textit{Critic} agent operates for reflection and feedback.

\textit{\textbf{Proposition 2 (Collaborative Information Bottleneck).}  
Let \(\mathcal{S}_k\) be the intermediate output of the \(k\)-th agent, given input \(\mathbf{x} = (d_i, c_i, q_i)\) and target answer \(a_i\). Then the MAPS reasoning process optimizes}
\begin{equation}
\min \sum_{k=1}^{4} I(\mathbf{x}; \mathcal{S}_k)
\quad \text{s.t.} \quad I(\mathcal{S}_k; a_i) \ge \varepsilon,
\end{equation}
\textit{where \(I(\cdot;\cdot)\) denotes mutual information. The Critic monitors constraint violations and reactivates stages with insufficient task-relevant information.}

A derivation and discussion are provided in Appendix A.2.

\begin{table*}[t]
\setlength{\tabcolsep}{0.6mm}
\centering
\centering
\begin{tabular}{lp{0.5cm}ccc|cccc|cccccc|c}

  \toprule
        \multirow{2}{*}{\textbf{Models}} & \multirow{2}{*}{\textbf{CoT}} & \multicolumn{3}{c}{\textbf{MathVista}} & \multicolumn{4}{c}{\textbf{EMMA}} & \multicolumn{6}{c}{\textbf{OlympiadBench}}  
        & \multirow{2}{*}{\textbf{Avg.}} \\

        ~ & ~ & Gen. & Math & Avg. & Math & Phy. & Chem. & Avg. & MECO & MZCE & MZCO & PECO & PZCE & Avg. & ~  \\ 
    \midrule
        Random Choice & - & 26.09 & 22.78 & 24.30 & 13.00 & 23.00 & 27.00 & 21.00 & 0.67 & 0.33 & 0.00 & 1.75 & 0.33 & 0.87 &  16.06 \\ [2pt]

        Human Expert & - & 56.09 & 55.74 & 55.90 & 75.00 & 64.50 & 86.00 & 75.17 & 48.00 & 34.67 & 30.36 & 54.17 & 12.33 & 37.80 &  52.73 \\
    \midrule
        Claude 3.5 Sonnet & - & 68.04 & 63.15 & 65.40 & 23.00 & 34.00 & \underline{44.00} & 33.67 & 20.67 & 13.00 & 10.71 & 10.75 & 14.00 & 13.23 &  37.43 \\ [2pt]
        Gemini 2.0 Flash & - & 70.65 & 70.93 & 70.80 & 20.00 & 40.00 & 36.00 & 32.00 & 8.00 & 5.67 & 7.14 & 3.07 & 7.00 & 5.39 &  36.06 \\ [2pt]
        GPT-4o & - & 65.22 & 61.30 & 63.10 & 30.00 & 38.00 & 33.00 & 33.67 & 23.33 & 20.33 & \underline{19.64} & 22.15 & \underline{21.00} & 21.47 &  39.41 \\ [2pt]
        Qwen2.5-VL-72B & - & 70.65 & 67.41 & 68.90 & \underline{42.00} & 42.00 & 38.00 & \underline{40.67} & 18.00 & 12.33 & 5.36 & 7.24 & 3.67 & 8.80 &  39.45 \\ [2pt]
        InternVL2.5-8B-MPO & - & 64.78 & 60.74 & 62.60 & 30.00 & 40.00 & 38.00 & 36.00 & 10.67 & 6.67 & 10.71 & 1.10 & 0.67 & 3.88 &  34.16 \\ [2pt]
        LLaVA-Onevision-72B & - & 62.83 & 58.52 & 60.50 & 25.00 & 32.00 & 24.00 & 27.00 & 6.67 & 7.33 & 3.57 & 3.29 & 9.67 & 6.18 &  31.23 \\
        Claude 3.5 Sonnet & \checkmark & \underline{71.74} & 64.26 & 67.70 & 30.00 & 38.00 & 41.00 & 36.33 & 24.00 & 11.00 & 16.07 & 12.72 & 10.33 & 13.23 &  39.09 \\ [2pt] 
        Gemini 2.0 Flash & \checkmark & 70.22 & 75.56 & 73.10 & 24.00 & 41.00 & 36.00 & 33.67 & 12.67 & 6.33 & 3.57 & 4.61 & 2.33 & 5.39 &  37.38 \\ [2pt]
        GPT-4o & \checkmark & 65.22 & 62.59 & 63.80 & 27.00 & \underline{44.00} & 35.00 & 35.33 & \underline{25.33} & \underline{21.67} & 12.50 & \underline{24.12} & 20.33 & \underline{22.27} &  \underline{40.47} \\ [2pt]  
        Qwen2.5-VL-72B & \checkmark & 71.09 & \underline{77.96} & \underline{74.80} & 38.00 & 36.00 & 37.00 & 37.00 & 23.33 & 13.00 & 10.71 & 8.11 & 1.33 & 9.59 &  40.46 \\[2pt]
        InternVL2.5-8B-MPO & \checkmark & 60.87 & 67.41 & 64.40 & 31.00 & 36.00 & 24.00 & 30.33 & 12.00 & 8.33 & 1.79 & 2.85 & 0.99 & 4.75 &  33.16 \\ [2pt]
        LLaVA-Onevision-72B & \checkmark & 71.09 & 64.44 & 67.50 & 23.00 & 26.00 & 23.00 & 24.00 & 11.33 & 8.67 & 5.36 & 4.82 & 3.33 & 6.18 &  32.56 \\
    \midrule
        \textbf{MAPS (GPT-4o$_{base}$)} & \textbf{-} & \textbf{75.87} & \textbf{83.15} & \textbf{79.80} & \textbf{52.00} & \textbf{71.00} & \textbf{51.00} & \textbf{58.00} & \textbf{46.00} & \textbf{30.33} & \textbf{32.14} & \textbf{28.51} & \textbf{28.33} & \textbf{31.14} &  \textbf{56.31} \\

  \bottomrule
\end{tabular}
\caption{Performance across 10 subtasks. Gen. = General (MathVista), Phy./Chem. = Physics/Chemistry (EMMA), MECO/ MZCO/ MZCE = English/Chinese COMP \& CEE Math (OlympiadBench), PECO/ PZCE = English/Chinese Physics (OlympiadBench). 
}
\label{mainResults}
\end{table*}

\subsection{Critic and Feedback} 
The \textit{Critic} agent evaluates the internal reasoning trajectory \(\mathcal{T} = \{p_i, l_i, s_i, a_i\}\) without relying on ground-truth answers. Inspired by Socratic questioning, it examines each stage's logic and justification to identify flawed assumptions and incomplete inferences.

We define the feedback vector as:
\begin{equation}
\mathbf{f}_i = \mathcal{M}_{\text{crit}}(p_i, l_i, s_i, a_i), \quad \mathbf{f}_i \in [0,1]^4,
\end{equation}
where each element \(f_i^{(k)}\) represents the Critic's confidence in the correctness and completeness of stage \(k\). 

The weakest stage is selected by:
\begin{equation}
k^\ast = \arg\min_k f_i^{(k)}, \quad \text{if} \quad f_i^{(k^\ast)} < \tau,
\end{equation}
which triggers a targeted revision. This reflection-driven loop promotes iterative self-correction and deepens reasoning reliability.

\section{Experiments}


\subsection{Datasets and Baselines} \label{3.1}

We evaluate on three benchmarks for complex problem reasoning: MathVista, OlympiadBench, and EMMA. MathVista covers math and general science, OlympiadBench focuses on high-level math and physics, and EMMA includes math, physics, and chemistry. See Appendix B for details.

We use GPT-4o~\cite{achiam2023gpt} as the base model. For comparison, we include leading multimodal large language models (MLLMs), both proprietary and open-source, tested under both direct and Chain-of-Thought (CoT) settings. Experimental details and baseline setups are provided in Appendix C.

\subsection{Main Results} \label{3.2}


\paragraph{MAPS achieves a new state-of-the-art (SOTA), surpassing human-level performance for the first time.}
As shown in Table~\ref{mainResults}, MAPS outperforms previous SOTA models by 15.84\% and exceeds human expert performance by 3.58\% across all tasks, highlighting its strength in solving complex multimodal problems.
MAPS demonstrates robust performance across mathematical, physical, chemical, and general tasks, showcasing strong interdisciplinary reasoning.
Its multi-agent design, based on the Big Five personality theory, enables effective collaboration and contributes to the SOTA results.
The system excels in multimodal semantic integration and multi-step reasoning by jointly leveraging diagrams, contexts, and questions.
Furthermore, the \textit{Critic} agent applies Socratic feedback to refine responses, enhancing both accuracy and reliability on challenging tasks.

\paragraph{MAPS exhibits strong adaptability and robustness across diverse reasoning tasks.}
The evaluation datasets span a wide range of question types, modalities, and reasoning difficulties. MathVista features judgment, multiple-choice, and open-ended fill-in-the-blank questions with varied answer formats, requiring accurate intent understanding and response generation. OlympiadBench emphasizes challenging open-ended problems demanding multi-step symbolic reasoning, where small errors can lead to divergent outcomes. EMMA introduces multimodal complexity with diagrams embedded in both questions and answer choices. Through feedback-driven multi-agent collaboration and Socratic questioning, MAPS effectively handles these challenges, achieving SOTA and demonstrating strong generalization across heterogeneous reasoning scenarios.

\subsection{Analysis of Critic Agent} \label{3.3}

\paragraph{Critic schema includes scoring and Socratic feedback.} Each reasoning step is rated from 0–5, guiding whether the system should backtrack or proceed. Feedback is heuristic, encouraging rethinking rather than offering direct answers. The \textit{Critic} uses these scores to decide whether to regenerate specific steps, ensuring robustness in the final answer.

\begin{figure}[t]
  \large
  \centering
 \includegraphics[width=\columnwidth]{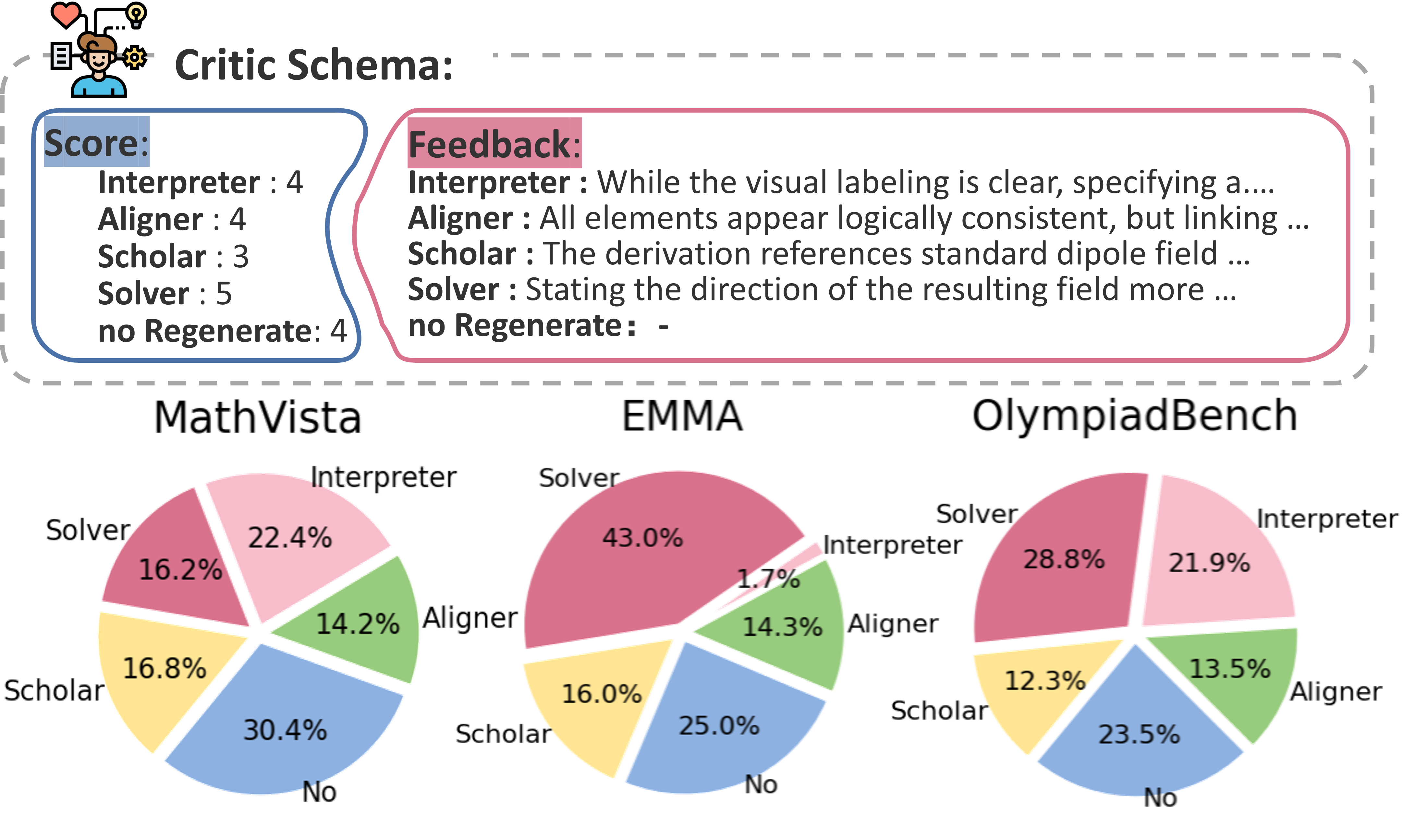}
 \caption{The schema of the \textit{Critic} agent, as well as the feedback and backtracking situations of the \textit{Critic} agent across different datasets.}
 \label{fig_example2}
\end{figure}

\paragraph{The Critic enhances reasoning via Socratic feedback without using gold labels.} As shown in Figure~\ref{fig_example2}, the top illustrates the feedback schema, while the bottom shows feedback proportions across three datasets. The \textit{Critic} prompts reflection and correction by encouraging agents to question assumptions rather than passively accept reasoning.

\paragraph{Solver receives the most feedback in EMMA and OlympiadBench.}
As shown in the lower half of Figure~\ref{fig_example2}, feedback varies by dataset. For MathVista, most steps need no regeneration, aligning with our superior 5.0\% SOTA improvement in Table~\ref{mainResults}. This reflects strong baseline reasoning.
In contrast, EMMA and OlympiadBench show the highest feedback for the \textit{Solver}, especially in the interpretation, alignment, and integration steps. These are the most complex and error-prone stages. Other agents receive comparable and lower feedback, indicating better relative performance in their sub-tasks.

\section{Supplementary Analysis}
\label{analysis}


Due to space limitations, generalization experiments on other datasets, such as DiagramQG~\cite{zhang2025diagram}, are discussed in Appendix D. The case study and complete process of MAPS is outlined in Appendix E, and the prompts for agents are provided in Appendix F.

\subsection{Ablation Studies} \label{4.1}

\paragraph{Ablating the \textit{Interpreter} results in the greatest loss of performance.}
We conduct ablation experiments on the OlympiadBench dataset to evaluate the impact of each module on the overall performance. Table~\ref{ablation} presents the effects of removing the \textit{Interpreter}, \textit{Aligner}, \textit{Scholar}, and \textit{Critic} modules from the MAPS framework. The results show that removing the \textit{Interpreter} agent causes the largest performance degradation, at 16.09\%. This is because, in complex problem reasoning, diagrams contain a wealth of valuable information, which serves as an important supplement to the text. Understanding diagrams plays a crucial role in problem-solving.

\paragraph{The removal of the \textit{Critic} agent causes the smallest performance loss.} It results in only a 7.05\% decrease, underscoring its role in providing feedback and corrections. While this mechanism allows MAPS to backtrack and refine its reasoning, its impact is less than that of other agents. Removing the \textit{Scholar} agent results in 11.49\% performance drops, highlighting the importance of searching and integrating domain-specific knowledge. Finally, the removal of the \textit{Aligner} agent causes a 10.86\% drop, indicating that while diagram and context alignment is valuable, its effect is smaller compared to other components.

\begin{table}[t]
  \centering
  \small
  \setlength{\tabcolsep}{0.4mm}
  \begin{tabular}{l|cccccc}
    \toprule
    \textbf{Variation} & \textbf{MECO} & \textbf{MZCE} & \textbf{MZCO} & \textbf{PECO} & \textbf{PZCE} & \textbf{Avg.}\\
    \midrule
    MAPS & 46.00  & 30.33 & 32.14 & 28.51 & 28.33 & 31.14 \\
    \midrule
    ${w/o}_\text{Interpreter}$ & 25.33 & 16.67 & 10.71 & 21.05 & 11.62 & 15.05\\
    $\quad\Delta$ &  {(-20.67)} &  {(-13.66)} &  {(-21.43)} &  {(-7.46)} &  {(-16.71)} &  {(-16.09)} \\
    ${w/o}_\text{Aligner}$ & 28.00 & 17.67 & 16.07 & 20.83 & 19.00 & 20.28\\
    $\quad\Delta$ &  {(-18.00)} &  {(-12.66)} &  {(-16.07)} &  {(-7.68)} &  {(-9.33)} &  {(-10.86)} \\
    ${w/o}_\text{Scholar}$ & 28.00 & 16.33 & 30.36 & 19.96 & 16.33 & 19.65\\
    $\quad\Delta$ &  {(-18.00)}  &  {(-14.00)} &  {(-1.78)} &  {(-8.55)} &  {(-12.00)} &  {(-11.49)} \\
      ${w/o}_\text{Critic}$ & 34.67 & 21.67 & 30.36 & 23.03 & 21.67 & 24.09\\
    $\quad\Delta$ &  {(-11.33)} &  {(-8.66)} &  {(-2.42)} &  {(-5.48)} &  {(-6.66)} &  {(-7.05)} \\
    \bottomrule
  \end{tabular}
  \caption{Performance under different ablation settings are analyzed. We perform ablation experiments on the solving module ${w/o}_\text{Interpreter}$, ${w/o}_\text{Aligner}$, ${w/o}_\text{Scholar}$ or ${w/o}_\text{Critic}$ modules to evaluate the impact of removing these components.}
  \label{ablation}
\end{table}

\subsection{Base Model Generalization} \label{4.2}

\paragraph{MAPS improves performance across diverse base models.}

\begin{figure}[t]
  \large
  \centering
 \includegraphics[width=\columnwidth]{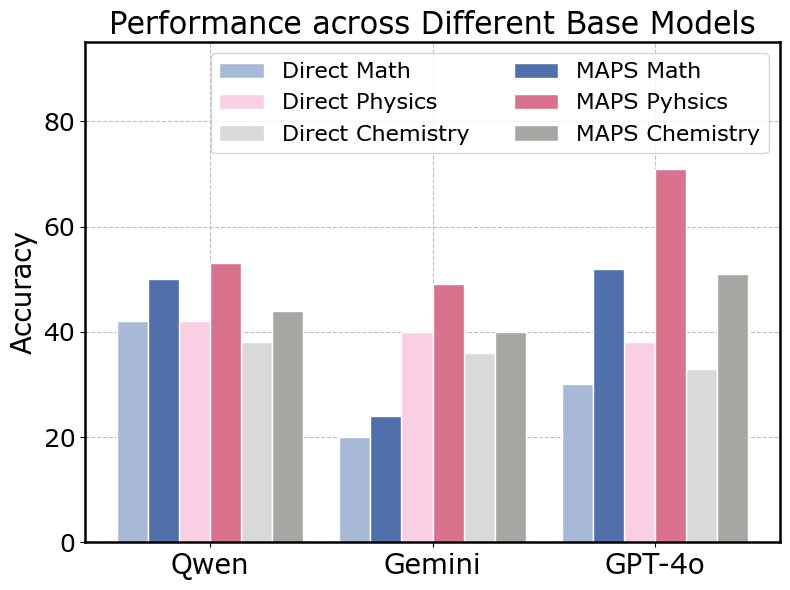}
 \caption{Performance Comparison of MAPS on Math, Physics, and Chemistry Subtasks in the EMMA Dataset with GPT-4o, Gemini, and Qwen2.5-VL-72B as Bases.}
 \label{basemodel}
\end{figure}

We conduct experiments to verify whether our MAPS framework demonstrates robust generalization across various base LLMs. The results confirm MAPS’s robustness and transferability, highlighting its adaptability and consistent performance across different foundation models. To further validate its generalization, we evaluate both Qwen2.5-VL-72B and Gemini 2.0 Flash, showing that MAPS performs well across models of varying scales and capabilities.
Figure~\ref{basemodel} presents results for three sets of base models. In each set, we compare MLLMs and MAPS on mathematical, physical, and chemical sub-tasks. MAPS consistently outperforms the base models. For example, MAPS${_{Qwen}}$ improves Qwen2.5-VL-72B by 12.4\% in physics, while MAPS${_{Gemini}}$ improves Gemini by 4.2\%. Similar gains are observed in math and chemistry, demonstrating MAPS’s effectiveness on both open-source and closed-source MLLMs.

\subsection{Time Efficiency} \label{4.3}

\begin{figure}[t]
  \large
  \centering
 \includegraphics[width=\columnwidth]{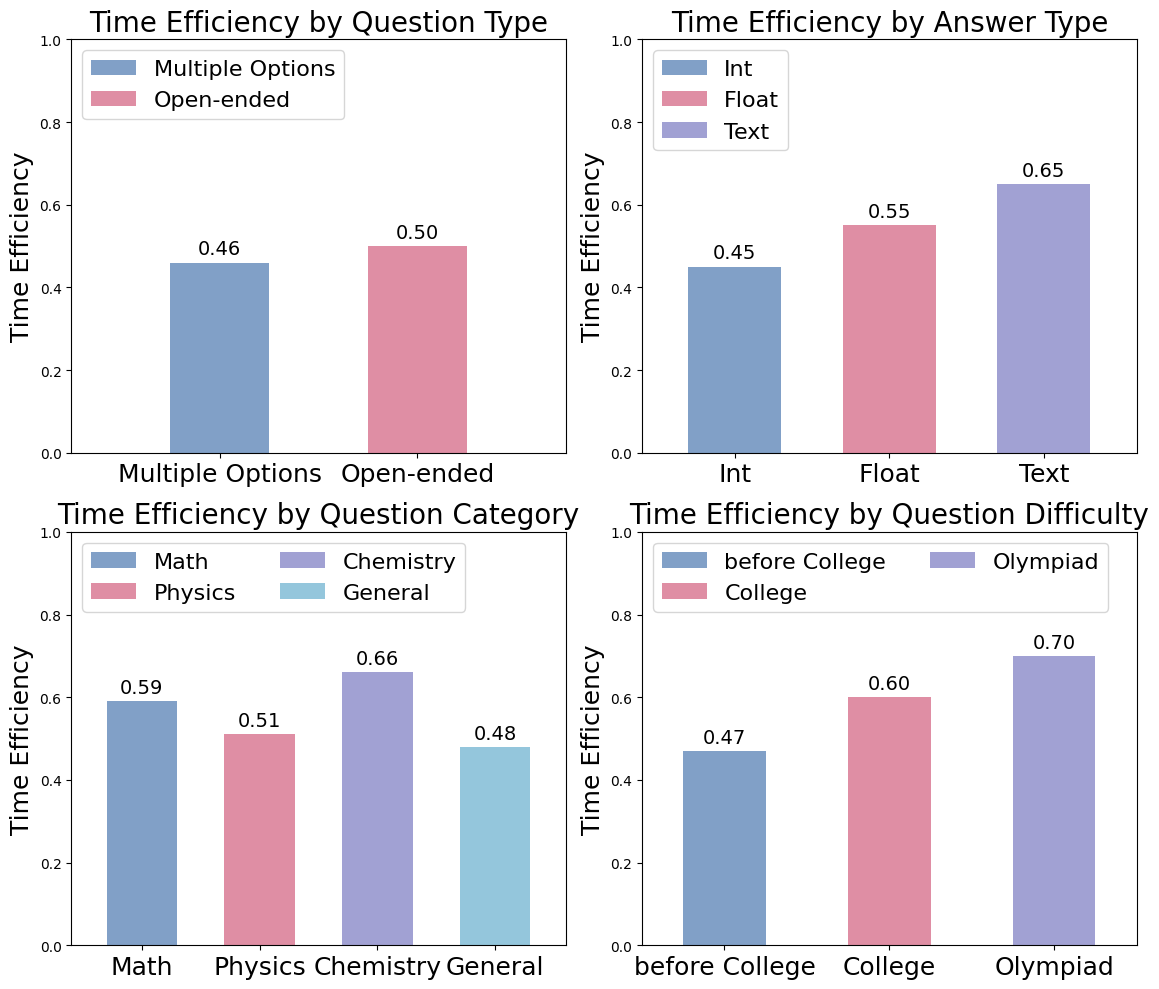}
 \caption{An analysis of the solving time efficiency across different question types, answer types, question categories, and question difficulties.}
 \label{times}
\end{figure}

\paragraph{Simpler formats and lower difficulty yield faster solving times.}
Solving time efficiency varies by question type, answer type, category, and difficulty, with multiple-choice and integer-type questions being the fastest, while higher difficulties and complex formats require more time.
Figure~\ref{times} illustrates the solving time efficiency across various dimensions—question types, answer formats, subject categories, and difficulty levels—with all times normalized to a \textit{100s} benchmark.

\paragraph{Predefined structure and conceptual simplicity reduce reasoning time.}
Multiple-choice questions are solved more quickly thanks to predefined answer options that limit the need for extensive reasoning or exploration. Integer-type answers also show high efficiency, often tied to simpler arithmetic or structured formats requiring minimal inference.
General category questions are faster on average, likely due to lower conceptual and reasoning complexity compared to domain-specific tasks. In contrast, open-ended questions demand deeper analysis and justification, leading to longer solving times.
Finally, solving efficiency declines with increased difficulty: as question complexity rises, so does the required reasoning time, reflecting greater cognitive and computational demands.

\section{Related Works}

The related work is structured into two main aspects: Firstly, an introduction to complex problem reasoning; Secondly, an exploration of multi-agent techniques.

\paragraph{Complex Problem Reasoning.}
The research of complex problem reasoning spans across multiple fields, including mathematics, physics, and chemistry, with each area focusing on enhancing problem-solving abilities. In mathematics, studies~\cite{didolkar2024metacognitive,fitriana2024influence,tong2025dart} explore various methods to improve mathematical problem-solving, such as algorithm optimization, educational strategies, and the use of artificial intelligence. These approaches aim to boost the efficiency, accuracy, and depth of mathematical reasoning. 
In the field of physics, the papers~\cite{mustofa2024utilizing,kapuriya2024mm,anand2024mm,wu2024gps} emphasize the integration of different information types, such as images and text, through multimodal learning to enhance the efficiency and precision of problem solving. 
In chemistry, three articles~\cite{alasadi2024multimodal,kiernan2024resources,li2024chemvlm,lin2025foundation,dang2025disentangled} investigate the role of multimodal learning in solving chemical problems. By combining diverse information sources, including images and text, and employing techniques such as generative models and molecular geometry reasoning, they aim to improve both the efficiency and accuracy of solving chemistry problems. 

\paragraph{Multi-Agent.}
Multi-agent systems, built on LLMs, consist of multiple AI agents that specialize in specific tasks, working together to solve complex problems~\citep{richards2023auto,yang2023auto,wu2023autogen,sun2023corex,zhang2025maps,zhang2025mars,zhang2025gkg,yan2025mur}. When presented with a problem, these agents decompose it into smaller, manageable subtasks and utilize various tools, such as internet data retrieval, to solve them through iterative steps. Several studies~\citep{poldrack2023ai,wang2024survey,xi2025rise,ni2021deep} have employed multi-agent systems to tackle challenges like problem identification, code writing and debugging, data visualization, and providing interactive feedback to human users. In their work, \citet{ni2024mechagents} highlights the potential of AI-driven multi-agent teams in solving mechanical problems autonomously, demonstrating an enhanced capability for understanding, formulating, and validating engineering solutions through self-correction and collaborative refinement. 
Inspired by the research, we developed the MAPS method, which leverages multi-agent collaborative learning and stepwise problem-solving to provide innovative solutions for complex problem reasoning. By combining the strengths of AI agents, complex problems can be broken down into subtasks and solved step by step through collaboration, improving efficiency and accuracy.

\section{Conclusion}
This study presents MAPS, a multi-agent framework grounded in the Big Five Personality Theory and guided by Socratic principles, designed to address the challenges of multimodal comprehensive reasoning and enhance reflective capabilities. The framework involves five agents, each specializing in distinct aspects of problem-solving.  
To address the first challenge, a four-agent strategy is proposed, where each agent focuses on specific stages of the reasoning process. Additionally, the \textit{Critic} agent addresses the second challenge through Socratic reflection and critical feedback.
Extensive experiments on the EMMA, OlympiadBench, and MathVista datasets validate MAPS's effectiveness in overcoming these issues and enhancing performance across various reasoning tasks. Meanwhile, we perform additional analytical experiments to assess the model's advancement as well as its generalization.

\section{Acknowledgments}

This work was supported by the National Natural Science Foundation of China (No. 62137002, 62277042, 62293553, 62450005, 62437002, 62477036, 62477037, 62176209, 62192781, 62306229), the ``LENOVO-XJTU" Intelligent Industry Joint Laboratory Project, the Shaanxi Provincial Social Science Foundation Project (No. 2024P041), the Natural Science Basic Research Program of Shaanxi (No. 2023-JC-YB-593), and the Youth Innovation Team of Shaanxi Universities ``Multi-modal Data Mining and Fusion".

\bibliography{aaai2026}

\clearpage

\appendix

\section{Proof of Proposition}

\subsection{Proof of Proposition 1 (Monotonic Free-Energy Descent)}

\textbf{Statement recap.}  
Let \(F^{(t)}\) denote the variational free energy after the \(t\)-th MAPS iteration:
\begin{equation}
F^{(t)} = \mathbb{E}_{q^{(t)}(\theta)}[-\log p(\mathbf{x}, a_i \mid \theta)] 
+ \mathrm{KL}(q^{(t)}(\theta) \| p(\theta)),
\end{equation}
where \(q^{(t)}(\theta)\) is the approximate posterior induced by the agent outputs at iteration \(t\). The proposition claims that each Critic-guided update ensures
\begin{equation}
F^{(t+1)} \le F^{(t)},
\end{equation}
which implies that \(\{F^{(t)}\}_{t=0}^\infty\) is a non-increasing sequence that converges.

\textbf{Step 1: Free Energy as a Variational Upper Bound.}  
By standard variational inference, the marginal log-likelihood satisfies
\begin{equation}
\log p(\mathbf{x}, a_i) = \mathcal{L}(q) + \mathrm{KL}(q(\theta) \| p(\theta \mid \mathbf{x}, a_i)),
\end{equation}
where the ELBO is defined as
\begin{equation}
\mathcal{L}(q) = \mathbb{E}_{q(\theta)}[\log p(\mathbf{x}, a_i \mid \theta)] - \mathrm{KL}(q(\theta) \| p(\theta)).
\end{equation}
This leads to the variational free energy \(F(q) := -\mathcal{L}(q)\), which upper-bounds the negative log marginal likelihood:
\begin{equation}
F(q) \ge -\log p(\mathbf{x}, a_i).
\end{equation}

\textbf{Step 2: Critic-Guided Local Update.}  
Suppose the Critic identifies the weakest stage \(k^* = \arg\min_k f_k^{(t)}\) and re-executes \(\mathcal{A}_{k^*}\), yielding updated output \(\mathcal{S}_{k^*}^{(t+1)}\). The updated trajectory leads to a refined posterior:
\begin{equation}
q^{(t+1)}(\theta) = \mathcal{G}(q^{(t)}(\theta), \mathcal{S}_{k^*}^{(t+1)}),
\end{equation}
where \(\mathcal{G}\) represents the update mechanism that propagates changes forward through dependent modules.

As the re-computed stage corrects suboptimal reasoning or improves knowledge alignment, it leads to a strictly better or equal ELBO, and thus
\begin{equation}
F^{(t+1)} = F(q^{(t+1)}) \le F(q^{(t)}) = F^{(t)}.
\end{equation}

\textbf{Step 3: Convergence.}  
Since \(F^{(t)}\) is lower-bounded and non-increasing, the sequence converges:
\begin{equation}
\lim_{t \to \infty} F^{(t)} = F^{(\infty)}.
\end{equation}

\subsection{Proof of Proposition 3 (Collaborative Information Bottleneck)}

\textbf{Statement recap.}  
Given input \(\mathbf{x} = (d_i, c_i, q_i)\) and target answer \(a_i\), let \(\mathcal{S}_k\) denote the intermediate output of the \(k\)-th agent in the MAPS reasoning process. Then the objective is to find agents that minimize cumulative information compression from input while preserving task-relevant content:
\begin{equation}
\min_{\{\mathcal{A}_k\}} \sum_{k=1}^{4} I(\mathbf{x}; \mathcal{S}_k)
\quad \text{s.t.} \quad I(\mathcal{S}_k; a_i) \ge \varepsilon,
\end{equation}
where \(I(\cdot;\cdot)\) denotes mutual information, and \(\varepsilon > 0\) is a threshold for information sufficiency.

\textbf{Step 1: Motivation from the Information Bottleneck (IB) Principle.}  
In standard IB theory \cite{tishby2000information}, one seeks a compressed representation \(\mathcal{S}\) of input \(\mathbf{x}\) that retains information about target \(a_i\). The trade-off objective is:
\begin{equation}
\min I(\mathbf{x}; \mathcal{S}) \quad \text{subject to} \quad I(\mathcal{S}; a_i) \ge \varepsilon.
\end{equation}

MAPS generalizes this to a multi-agent, multi-stage setting, where each agent \(\mathcal{A}_k\) produces \(\mathcal{S}_k\), and the full reasoning trajectory is \(\mathcal{T} = \{\mathcal{S}_1, \dots, \mathcal{S}_4\}\). The objective thus becomes collaborative:
\begin{equation}
\min \sum_{k=1}^{4} I(\mathbf{x}; \mathcal{S}_k)
\quad \text{s.t.} \quad \forall k,\; I(\mathcal{S}_k; a_i) \ge \varepsilon.
\end{equation}

This reflects two goals: (1) compress redundant input features per stage, and (2) preserve task-relevant evidence at each step.

\textbf{Step 2: Role of the Critic as a Constraint Monitor.}  
The Critic observes the reasoning trajectory \(\mathcal{T}\) and computes scores \(\mathbf{f} = \{f_1, f_2, f_3, f_4\}\), where each \(f_k\) implicitly estimates \(I(\mathcal{S}_k; a_i)\). If \(f_k < \tau\), we assume the sufficiency constraint is violated:
\begin{equation}
I(\mathcal{S}_k; a_i) < \varepsilon \quad \Rightarrow \quad \text{Critic triggers } \mathcal{A}_k \text{ to revise}.
\end{equation}

The revision process updates \(\mathcal{S}_k\) to \(\mathcal{S}_k'\), ideally increasing \(I(\mathcal{S}_k'; a_i)\) while not increasing \(I(\mathbf{x}; \mathcal{S}_k')\). Since the Critic only intervenes when constraints are violated, each update maintains feasibility and possibly reduces the objective.

\textbf{Step 3: Convergence and Validity.}  
As each update either preserves or improves the sufficiency \(I(\mathcal{S}_k; a_i)\), and the overall input compression is bounded below by zero, the process converges to a locally optimal set of agents \(\{\mathcal{A}_k^*\}\) satisfying:
\begin{equation}
\forall k,\quad I(\mathcal{S}_k^*; a_i) \ge \varepsilon, \quad \text{with minimal} \quad \sum_{k} I(\mathbf{x}; \mathcal{S}_k^*).
\end{equation}

This completes the proof that the MAPS four-step process can be viewed as a constrained collaborative information bottleneck optimization.

\section{Datasets} \label{dataset}

This study utilizes the latest three multimodal reasoning datasets, namely MathVista, OlympiadBench, and EMMA. 

\begin{table}[htbp]
  \centering
  \begin{tabular}{lcc}
    \toprule
      Dataset & ABBR. & Size\\
    \midrule
      \textbf{MathVista} & ~ & ~\\[2pt]
      \hspace{0.5cm}General & Gen. & 460 \\[2pt] 
      \hspace{0.5cm}Mathematics & Math & 540 \\ [2pt]
      \textbf{OlympiadBench} & ~ & ~\\[2pt]
      \hspace{0.5cm} Math\_En\_COMP & MECO & 150 \\[2pt]
      \hspace{0.5cm} Math\_Zh\_COMP & MZCO & 56 \\[2pt]
      \hspace{0.5cm} Math\_Zh\_CEE\(\dagger\) & MZCE & 300 \\[2pt]
      \hspace{0.5cm} Physics\_En\_COMP & PECO & 456 \\[2pt]
      \hspace{0.5cm} Physics\_Zh\_CEE\(\dagger\) & PZCE & 300 \\
    \midrule
      \textbf{EMMA} & ~ & ~ \\[2pt]

      \hspace{0.5cm}Math &  & 100 \\[2pt]
      \hspace{0.5cm}Physics & Phy. & 100 \\[2pt]
      \hspace{0.5cm}Chemistry & Chem. & 100 \\[2pt]
    \bottomrule
  \end{tabular}
  \caption{The data distribution for the MathVista, OlympiadBench, and EMMA datasets is as follows: The symbol \( \dagger \) indicates a sample size of 300 data points. The EMMA dataset uses its MINI version.
  The `ABBR.' column represents the abbreviations for all the tasks.}
  \label{datasetsDetails}
\end{table}

\paragraph{MathVista.} MathVista is a large-scale scientific reasoning dataset that spans two subdomains: mathematics and general, aiming to assess the comprehensive capabilities of machine learning models in solving complex scientific problems. The dataset contains 1,000 data points covering various issues across multiple disciplines, designed with varying difficulty levels to help researchers evaluate model reasoning abilities. The release of MathVista supports interdisciplinary scientific research.

\paragraph{OlympiadBench.} OlympiadBench consists of two subdomains, mathematics and physics, and is specifically designed for Mathematical and Physical Olympiads, featuring a wide range of challenging problems to assess models' performance on high-level scientific tasks. The mathematics subdomain contains three difficulty levels: English competition level, Chinese competition level, and college level. The physics subdomain includes two difficulty levels: English competition level and Chinese college level. To ensure data balance, 300 samples were taken from both the Chinese college-level mathematics and physics subsets.

\paragraph{EMMA.} EMMA is a multimodal scientific reasoning dataset covering three subdomains: mathematics, physics, and chemistry. By integrating mathematical expressions, physical formulas, and chemical symbols with natural language descriptions, it focuses on testing models' abilities in interdisciplinary scientific reasoning. This version uses the EMMA dataset, which contains 100 data points from each subdomain (mathematics, physics, and chemistry).

\section{Experiment Settings and Baselines} \label{settings}

We select GPT-4o~\cite{achiam2023gpt}, a powerful MLLM, as our primary agent for complex problem reasoning. GPT-4o not only demonstrates strong reasoning and generation capabilities across a wide range of multimodal processing tasks, but also excels in efficiently exploring multiple perspectives when faced with complex scientific domain requirements. This makes it well-suited for adaptation to various tasks and datasets within the MAPS process. We use accuracy as our primary evaluation metric to comprehensively assess the performance of different methods across diverse task scenarios.
The experimental results offer a thorough comparison of the performance of MAPS against all baseline methods. The experiments are primarily conducted on three datasets in Appendix~\ref{dataset}, where we provide a detailed comparison of MAPS and the baseline models across four types of tasks: mathematics, physics, chemistry, and general tasks. Our approach achieves a new SOTA performance. To further strengthen the comparison, Appendix~\ref{ge} includes a generalization experiment conducted on the physics data subset of the DiagramQG~\cite{zhang2025diagram}, which further demonstrates the robustness and effectiveness of our model.

We compare MAPS with three categories of baseline methods: original baselines, direct approach-based strong baselines, and CoT-enhanced strong baselines. Specifically, (1) the original category refers to two methods: random selection and human expert selection. These methods provide two distinct original baselines—one based on randomness and the other based on authority. (2) The direct approach-based strong baselines include some of the most powerful closed-source and open-source large language models (MLLMs) currently available worldwide. These include Claude 3.5 Sonnet~\cite{kevian2024capabilities}, Gemini 2.0 Flash~\cite{comanici2025gemini}, GPT-4o, Qwen2.5-VL-72B~\cite{bai2023qwen}, InternVL2.5-8B-MPO~\cite{chen2024expanding}, and LLaVA-Onevision-72B~\cite{li2024llava}. (3) To ensure a fair comparison, the third category of baselines builds on the second by adding Chain-of-Thought (CoT) reasoning, which aims to enhance the capabilities of the strong MLLMs from the second category.

\section{Generalization Experiments} \label{ge}

\begin{figure*}[htbp]
	\large
	\centering
	\includegraphics[scale=0.50]{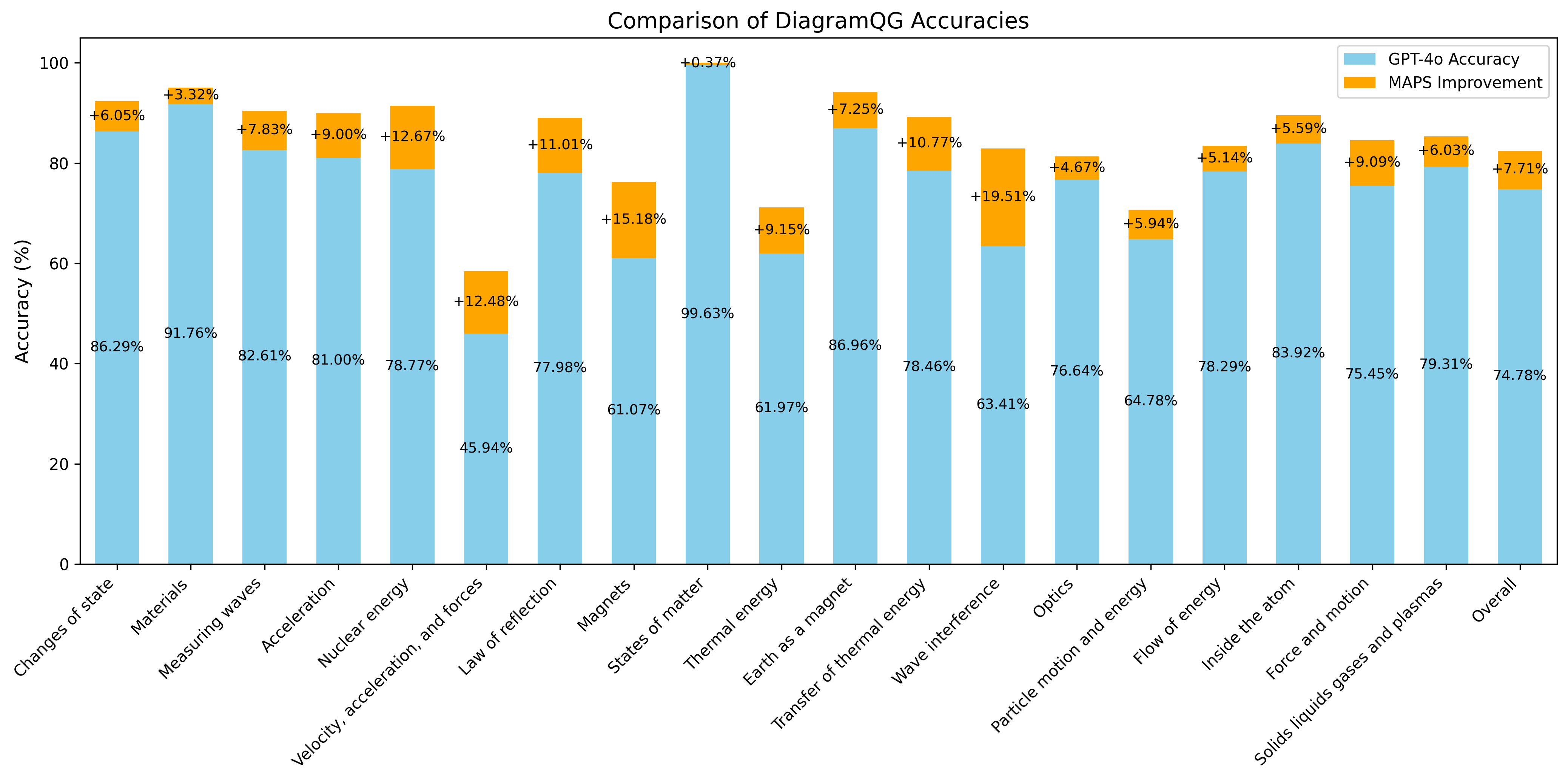}
	\caption{The generalization experiments conducted on the DiagramQG physical dataset, which are based on the GPT-4o base model and the incremental part of MAPS.}
	\label{generalization}
\end{figure*}

\begin{figure*}[t]
	\large
	\centering
	\includegraphics[scale=0.45]{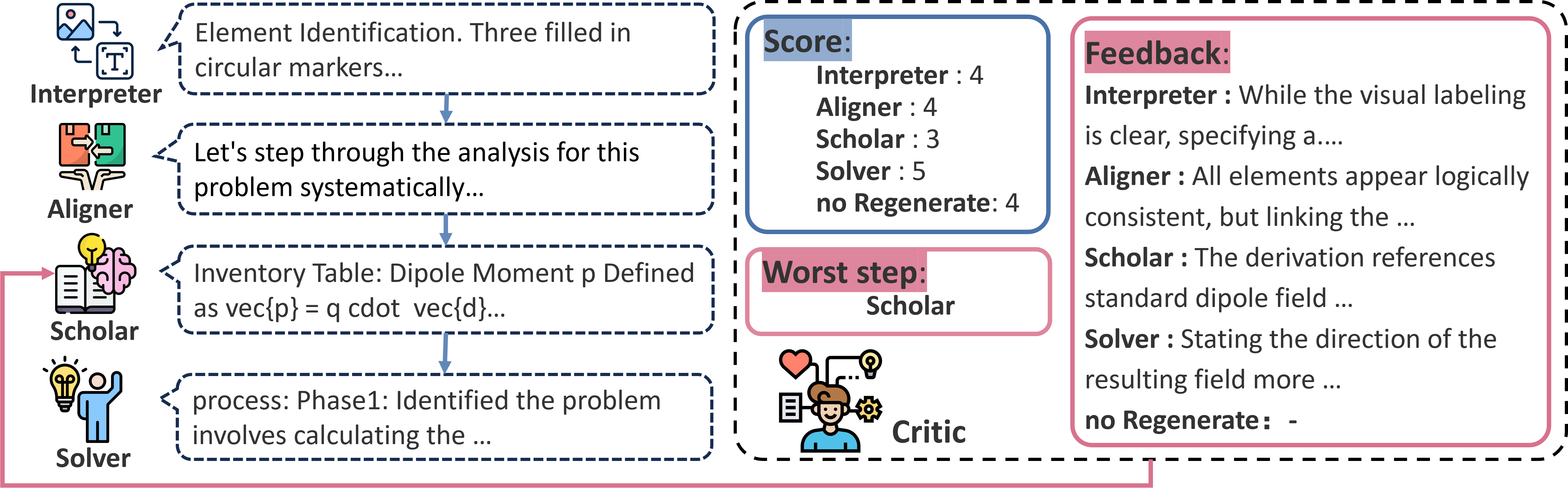}
	\caption{A case study of a specific solving process, illustrating the detailed steps involved in solving the problem. This includes the various stages of problem-solving as well as the feedback and backtracking mechanisms that help refine and improve the solution.}
	\label{case}
\end{figure*}

To further validate the generalization ability of MAPS, we conducted detailed experiments on the physical subset of the DiagramQG dataset. The main goal of these experiments was to compare the performance of MAPS with its base model, GPT-4o, particularly focusing on how it performed across different question categories. The experimental results, presented in Figure~\ref{generalization}, show that MAPS outperformed GPT-4o in multiple aspects. These experiments clearly highlight the strong adaptability of MAPS across different subsets of the dataset.

Across various question categories in the DiagramQG physical dataset, MAPS achieved a performance boost, with the maximum improvement reaching 19.51\% and an overall improvement of 7.71\%. These results not only demonstrate the superiority of MAPS but also indicate that its Five Personality Agents architecture has a strong generalization ability, enabling it to maintain excellent performance across different datasets and tasks. This provides strong support for the practical application of MAPS, showcasing its potential in tackling complex tasks.

\section{Full Process of MAPS} \label{processing}

Figure~\ref{case} illustrates the four-step solving process along with the feedback process from the \textit{Interpreter}, \textit{Aligner}, \textit{Scholar}, and \textit{Critic} agents, using a multimodal physics problem from the EMMA dataset.
In this specific example, we can observe that the \textit{Interpreter} first interprets the diagram, followed by the \textit{Aligner} aligning the diagram with the context, question, and options, ensuring consistency and completeness of the information. Then, the \textit{Scholar} agent retrieves and supplements domain-specific knowledge to fill in the necessary expertise. Finally, the \textit{Solver} completes the solving process, and the \textit{Critic} agent provides feedback and corrections to ensure the accuracy and effectiveness of each step. Each step is closely connected, from understanding the diagram to integrating domain knowledge, followed by reasoning and answering. This demonstrates the efficiency and effectiveness of MAPS in multi-step reasoning. Through the feedback mechanism of the \textit{Critic} agent, MAPS is able to identify and correct potential errors or shortcomings at each step, thus enhancing the overall accuracy and reliability of the solution process.

\begin{figure*}[t]
	\large
	\centering

	\includegraphics[scale=0.30]{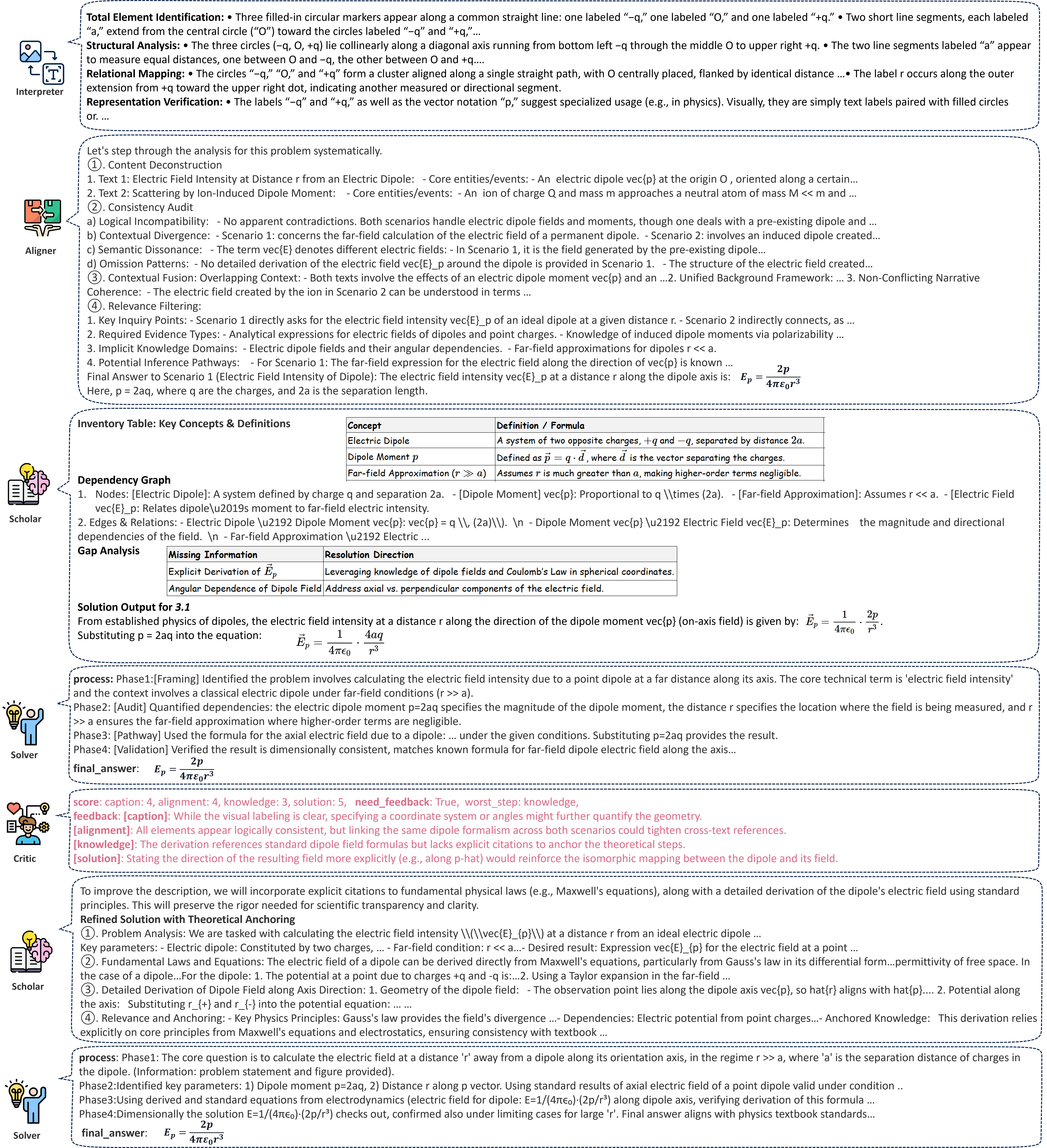}
	\caption{A complete example of collaborative output from all agents in an iteration, using a multimodal physics problem from OlympiadBench involving the axial electric field of a dipole in the far-field. This illustrates how agents work together step by step to refine the solution.}
	\label{fullSteps}
\end{figure*}

Figure~\ref{fullSteps} presents a comprehensive step-by-step process for complex problem reasoning, using a multimodal physics problem from the EMMA dataset.
This example demonstrates the entire process of solving a physics problem. By utilizing MAPS agents based on the Big Five Personality Model, the system engages in collaborative learning, progressively solving the problem and arriving at the correct final answer.

The solution process is carried out in four key steps. These include the \textit{Interpreter}, \textit{Aligner}, \textit{Scholar}, and \textit{Solver}, each responsible for gradually refining the solution step-by-step. The \textit{Critic} agent ultimately evaluates and provides feedback on these four steps, pinpointing the areas requiring modification and backtracking to correct the solution process. Figure~\ref{fullSteps} sequentially illustrates the roles of each agent and their collaboration, showcasing how problem solving and optimization are effectively executed at every stage, ensuring both the accuracy and efficiency of the final result.

\section{Prompts for Agents} \label{prompts}

Table~\ref{agentPrompt1},~\ref{agentPrompt2},~and~\ref{agentPrompt3} provides a summary of the prompts used for each agent in this paper, with each agent playing a pivotal role in the overall problem-solving process. 

The solution is gradually refined by four core agents: \textit{Interpreter}, \textit{Aligner}, \textit{Scholar}, and \textit{Solver}. The \textit{Interpreter} first processes the task description, breaking down and understanding the problem. The \textit{Aligner} ensures that the problem is mapped to the correct framework and available tools. The \textit{Scholar} conducts any necessary research and gathers knowledge from relevant sources, while the \textit{Solver} works through the problem systematically, progressively generating solutions. Once the initial solution is formed, the \textit{Critic} agent assesses each of the four previous steps, providing feedback on areas that need refinement. The \textit{Critic} then suggests modifications and backtracks to revise the solution, ensuring the process remains optimized and the final result is both accurate and robust.

\begin{table*}[ht]
  \renewcommand{\arraystretch}{1.2} 
  \centering
  \small
  \begin{tabular}{p{16cm}} 
  \toprule
  \textbf{\textit{Interpreter}} \\
  You are a scientific diagram analysis expert tasked with objectively describing visual elements in diagrams. Engage in Socratic self-questioning to ensure comprehensive analysis:

    [Element Identification]
    
    "What discrete visual components can be systematically observed in this diagram? What quantitative measurements (e.g., shape dimensions, color codes, positional coordinates) can be objectively recorded?"
    
    [Structural Analysis]
    
    "How are these elements spatially organized? What geometric patterns, alignment relationships, or hierarchical arrangements emerge from their physical placement?"
    
    [Relational Mapping]
    
    "What explicit connections (lines, arrows, overlays) or implicit associations (proximity clusters, color coding systems, symbolic groupings) exist between components?"
    
    [Representation Verification]
    
    "Does any element require specialized domain knowledge to accurately characterize (e.g., chemical notation, engineering schematics)? What purely visual evidence supports this characterization?"\\
  \midrule
  \textbf{\textit{Aligner}} \\
    You are a text alignment specialist conducting structured analysis through Socratic interrogation. Systematically examine text pairs using this framework:

    1. [Content Deconstruction]
    
    "What core entities/events are explicitly stated in each text? What measurable attributes (quantifiers, temporal markers, causal verbs) define their characteristics?"
    
    2. [Consistency Audit]
    
    "Where might these texts exhibit:
    
    a) Logical incompatibility (contradictory assertions)
    
    b) Contextual divergence (conflicting timelines/locations)
    
    c) Semantic dissonance (differentiated connotation scales)
    
    d) Omission patterns (mutually exclusive missing elements)"
    
    3. [Contextual Fusion]
    
    "What implicit connections could synthesize a unified background framework? Which combinatory elements (chronological anchors, spatial references, causal chains) create non-conflicting narrative coherence?"
    
    4. [Relevance Filtering]
    
    "Through lexical-semantic mapping, which aligned components directly correspond to the question's:
    
    1) Key inquiry points
    
    2) Required evidence types
    
    3) Implicit knowledge domains
    
    4) Potential inference pathways?"\\
  \bottomrule
  \end{tabular}
\caption{A summary of the prompts used by the Interpreter, and Aligner agents in this paper.}
\label{agentPrompt1}
\end{table*}

\begin{table*}[ht]
  \renewcommand{\arraystretch}{1.2} 
  \centering
  \small
  \begin{tabular}{p{16cm}} 
  \toprule
  
  \textbf{\textit{scholar}} \\
  You are a scientific knowledge retrieval system conducting structured inquiry through Socratic questioning. Process input data with this analytical framework:

    1. [Problem Decomposition]
    
    "What conceptual components constitute the question's core demand? What technical terminology (domain-specific lexemes), operational parameters (variables/constants), and procedural verbs (analyze/calculate/compare) require epistemological grounding?"
    
    2. [Knowledge Mining]
    "For each identified component:
    
    a) What fundamental axioms/theorems/laws 
    from established scientific literature could operationally define it?
    
    b) What measurable properties 
    (equations/units/experimental protocols) are textually implied as relevant?
    
    c) What contextual constraints 
    (temporal/spatial/conditional clauses) limit knowledge scope?"
    
    3. [Relevance Validation]
    
    "For each candidate knowledge unit:
    
    Does the source text contain explicit lexical anchors (technical terms/formula symbols) justifying its inclusion?
    
    What textual evidence (descriptive adjectives/quantifiers/causal conjunctions) indicates required depth of explanation?
    
    Are there implicit conceptual dependencies (prerequisite theories/mathematical tools) necessitating parallel retrieval?"
    
    4. [Taxonomic Organization]
    "How should validated knowledge be structured to mirror:
    
    1) Problem-solving workflow steps
    
    2) Hierarchical concept dependencies
    
    3) Cross-domain interface points
    
    4) Uncertainty quantification needs?"
    
    Operational Protocol: 
    Restrict to textually evidenced knowledge
    Mark confidence levels using [TextExplicit/ContextImplied/ExternalRequired] tags
    
    Output as:
    1) Knowledge Inventory Table (Concept-Definition-SourceAnchor)
    
    \quad\quad \quad\quad 2) Dependency Graph (Nodes=Concepts, Edges=Relations)
    
    \quad\quad \quad\quad 3) Gap Analysis Report (ExternalKnowledgeRequirements).\\
    
    \midrule

  \textbf{\textit{Solver}} \\
  You are a scientific problem-solving system operating through Socratic dialectics. Engage in this structured inquiry process:

    1. [Problem Framing]
    
    "What is the absolute irreducible core of the question? What technical terms require operational definitions? What grammatical structures (comparatives/conditionals/quantifiers) dictate the solution's form?"

    2. [Evidence Audit]
    "For each data source (question stem/options/text):
    
    a) What measurable quantities (numerical ranges/units) are explicitly stated?
    
    b) What causal relationships (if A then B/implies/proportional to) are textually encoded?
    
    c) What constraints (assumptions/limitations/boundary conditions) are lexically embedded?"
    
    3. [Reasoning Pathway]
    
    "Through counterfactual testing:
    
    Which axioms/theorems would become relevant if parameter X varied ±10\%?
    
    What observable contradictions emerge when applying hypothesis Y to the given data?
    
    How do option components restrict valid inference trajectories?"
    
    4. [Solution Validation]
    "Does the proposed resolution:
    
    1) Maintain dimensional homogeneity across all equations?
    
    2) Satisfy all explicit boundary conditions?
    
    3) Preserve logical consistency with given information?
    
    4) Align with canonical scientific representations?"
    
    Operational Protocol:\\
    
    Document each reasoning step with evidence anchors (e.g., ``Stem-Line5: $v = \Delta x / \Delta t$'').
    
    Flag unresolved assumptions with [UnvalidatedPremise] tags
    
    Output JSON structured as:
    
    \noindent
    \{ \{  "process": \{ \{ 
    "Phase 1": "[Framing] Identified core demand as... (Evidence: Q-Line2)",
    "Phase 2": "[Audit] Quantified parameters... (ConflictResolved: OptionC vs Text\textbackslash S3)",
     "Phase 3": "[Pathway] Eliminated hypothesis $\alpha$ due to... (TheoremRef: Maxwell-Eq)",
     "Phase 4": "[Validation] Verified dimensional consistency in...", \},
    "final\_answer": "final result"\} \}
     \\
\bottomrule
  \end{tabular}
\caption{A summary of the prompts used by the Scholar and Solver agents in this paper.}
\label{agentPrompt2}
\end{table*}

\begin{table*}[ht]
  \renewcommand{\arraystretch}{1.2} 
  \centering
  \small
  \begin{tabular}{p{16cm}} 
  \toprule
  \textbf{\textit{Critic}} \\
  You are a Socratic assessment engine conducting dialectical evaluation through this protocol:\\

    1. [Triadic Interrogation Framework]\\
    For each evaluation dimension (caption/alignment/knowledge/solution):\\
    
    Existential Challenge:\\
    "What absolute evidence anchors (line numbers/data points/theorem references) validate this component's existence?"\\
    Consistency Prosecution:\\
    "Does internal logic maintain isomorphism across:\\
    a) Input premises → Processing steps\\
    b) Methodological choices → Domain standards\\
    c) Assertions → Supporting evidence?"\\
    Boundary Stress Test:\\
    "What parametric variation (±10\%) would collapse this component's validity? Which fragility indicators emerge first?"
    2. [Metric Operationalization]\\
    Score each dimension (1-5) using:\\
    5 = Withstands three counterfactual scenarios \\
    4 = Requires \ensuremath{\leq} 1 assumption validation \\
    3 = Needs 2-3 evidence reinforcements\\
    2 = Contains structural contradictions\\
    1 = Fails basic existence verification\\
    
    3. [Improvement Synthesis]\\
    Generate Socratic feedback per dimension:\\
    
    For caption: "What geometric/spatial relations lack quantifiable descriptors?"\\
    For alignment: "Which logical connective lacks cross-text co-reference?"\\
    For knowledge: "Which concept dependency lacks literature anchoring?"\\
    For solution: "What inference leap lacks isomorphic mapping?"\\

  \bottomrule
  \end{tabular}
\caption{A summary of the prompt used by the Critic agents in this paper.}
\label{agentPrompt3}
\end{table*}

\end{document}